\documentclass[letterpaper, 10 pt, conference]{ieeeconf}  

\IEEEoverridecommandlockouts                              

\newcommand{\best}[1]{{\setlength{\fboxsep}{1pt}\colorbox{black!16}{\textbf{#1}}}}
\newcommand{\second}[1]{\underline{#1}}
\newcommand{\tbd}{--}

\usepackage{graphicx}
\usepackage{amsmath}
\usepackage{amssymb}
\usepackage{cite}
\usepackage{blindtext}
\usepackage[table]{xcolor}
\usepackage{multirow}
\usepackage{rotating}
\usepackage{booktabs}

\newcommand{\PAR}[1]{\vskip 0pt \noindent{\bf #1~}}
\makeatletter
\def\subsection{\@startsection{subsection}{2}{\z@}{0.6ex plus 0.6ex minus 0.3ex}%
{0.3ex plus .2ex minus 0ex}{\normalfont\normalsize\itshape}}%
\makeatother

\makeatletter
\long\def\@makecaption#1#2{%
\@IEEEfigurecaptionsepspace%
\setbox\@tempboxa\hbox{\footnotesize #1:~ #2}%
\ifdim \wd\@tempboxa >\hsize%
\setbox\@tempboxa\hbox{\footnotesize #1:~ }%
\parbox[t]{\hsize}{\footnotesize \noindent\unhbox\@tempboxa#2}%
\else%
\ifcenterfigcaptions \hbox to\hsize{\footnotesize\hfil\box\@tempboxa\hfil}%
\else \hbox to\hsize{\footnotesize\box\@tempboxa\hfil}%
\fi\fi%
\vspace{3pt}}
\makeatother

\title{\LARGE \bf
Polygraph: Large-Scale Multi-Session Trajectory Optimization using Pose Graphs
}

\author{Petar Lukovic$^{1}$, Tian Yi Lim, Marc Pollefeys$^{1,2}$, Pablo Speciale$^{2}$, Hermann Blum$^{3}$, Zuria Bauer$^{1}$
\thanks{$^{1}$ETH Zurich, Switzerland.}%
\thanks{$^{2}$Microsoft Spatial AI Lab, Zurich, Switzerland.}%
\thanks{$^{3}$University of Bonn and Lamarr Institute for Machine Learning and Artificial Intelligence, Germany.}%
\thanks{Corresponding author: Petar Lukovic, {\tt\small plukovic@ethz.ch}}%
}

\makeatletter
\renewcommand{\fnum@table}{Table~\thetable}
\makeatother

\begin{document}

\maketitle
\thispagestyle{empty}
\pagestyle{empty}

\begin{abstract}

Robots, vehicles, and wearable devices often traverse the same environment many times. Each pass produces its own trajectory from onboard visual or visual-inertial odometry. These trajectories are locally accurate but drift over time and live in unrelated coordinate frames. Turning many such independent trajectories into one globally consistent set is the large-scale multi-session trajectory optimization problem. Existing solutions largely work offline. All sessions must be available up front, one session is fixed as the reference frame, and a single fused map is then optimized. This paper presents Polygraph, a multi-session trajectory optimization pipeline that processes sessions online, one keyframe and one spatial cluster at a time, and never builds a 3D map. Keyframes are grouped into local clusters. When a cluster closes, it is matched against every other session through learned image retrieval. Candidate pairs are verified with sparse feature matching, monocular metric depth, and robust pose estimation. A two-tier consensus gate admits a cross-session edge into the shared pose graph only once independent evidence corroborates it. A cheap incremental solve keeps the graph current, and a periodic robust batch solve prunes outliers. We evaluate Polygraph on stereo outdoor driving, HoloLens and iPad indoor traversals, legged indoor locomotion with a Boston Dynamics Spot, and large-scale outdoor vehicle and handheld routes. Against the closest prior work on the same GPU model, using its reported timings, Polygraph runs faster, commits fewer loop-closure edges, and matches or improves on accuracy.

\end{abstract}

\section{INTRODUCTION}

\begin{figure}[t]
  \centering
  \includegraphics[width=\columnwidth]{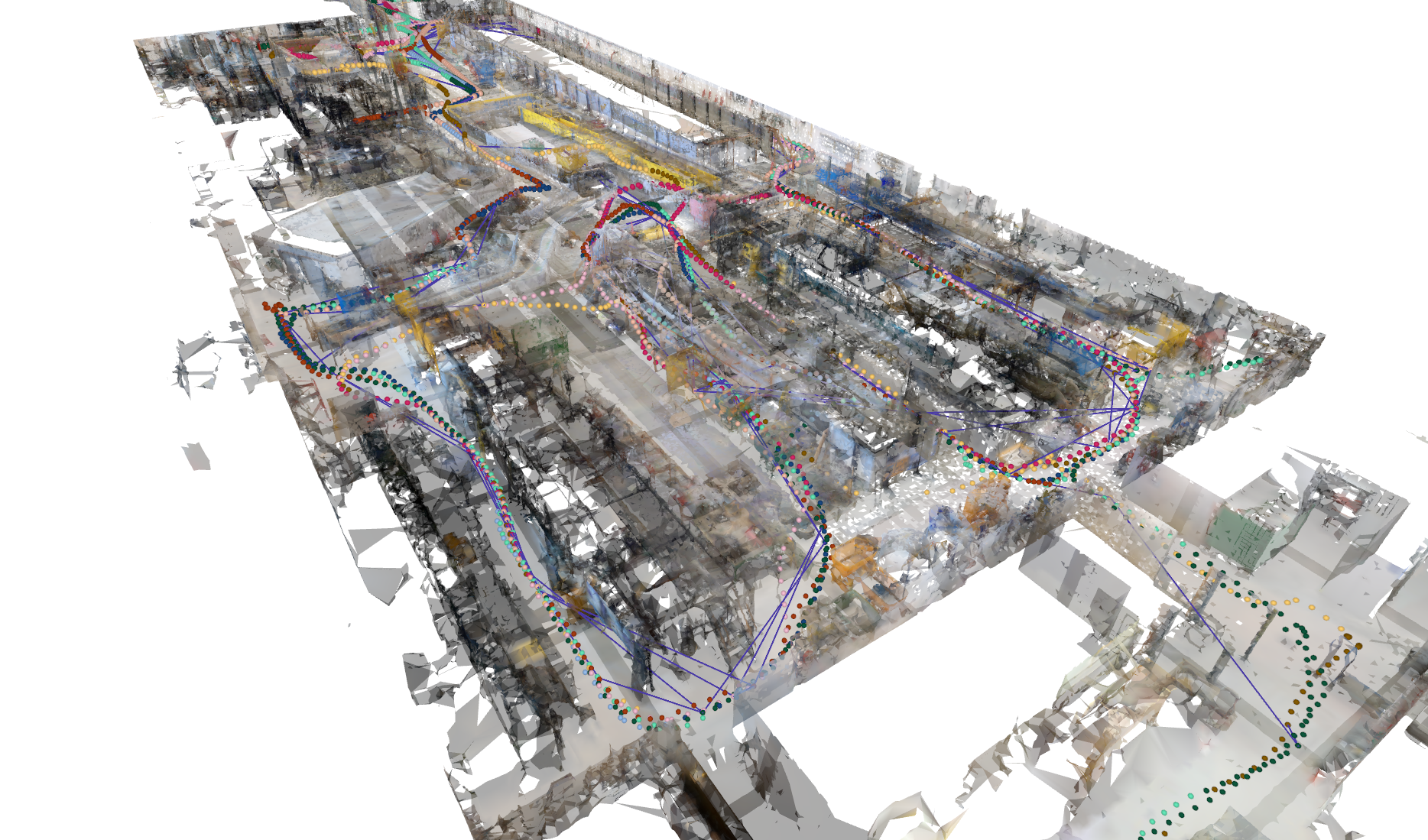}
  \caption{\textbf{Loop-closure visualisation}. Five independent HYDRO sessions recorded with a Boston Dynamics Spot, each in its own color, placed into one shared frame by Polygraph and overlaid on the site's 3D scan. Purple lines mark confirmed cross-session loop-closure edges. No two of these trajectories start out knowing that they share the same building.}
  \label{fig:teaser}
\end{figure}

Many robots, vehicles, and wearable devices see the same environment more than once. A delivery robot drives the same warehouse floor every day. A person wearing an AR headset walks through the same building on many occasions. We call each such pass a session. Each session is tracked by the device's own visual or visual-inertial odometry. This odometry is accurate locally but drifts over time, and every session lives in its own coordinate frame. Two sessions that walk the same corridor have no way of knowing that they did. Recovering that relationship is the multi-session trajectory optimization problem. It means detecting where sessions overlap, using those overlaps to correct drift, and placing every session into one shared frame. Fig.~\ref{fig:teaser} shows what this looks like once solved.

The classical way to detect overlap is to build and maintain a 3D map of visual landmarks and match new observations against it, as in ORB-SLAM3 Atlas~\cite{elvira2019orbslamatlasrobustaccuratemultimap}, Maplab~\cite{schneider2018maplab,cramariuc2023maplab2}, and COVINS~\cite{covins}. A map gives strong geometric evidence for a match, but is expensive to store and keep optimized; bundle adjustment grows linearly in landmarks and cubically in poses, so this cost grows with every session and with the size of the environment. A well-built map is also rarely what downstream modules want, since planning and navigation seldom consume a landmark cloud directly.

A recent line of work avoids keeping any map at all. Multi-robot systems such as DOOR-SLAM~\cite{lajoie2020doorslam}, Swarm-SLAM~\cite{lajoie2024swarmslam}, and FLOCC-SLAM~\cite{lajoie2025flocc} treat closures as pose-graph edges directly but still assume every robot is known up front. 2GO~\cite{lim2025twogo} instead represents a single session only as a sparse keyframed pose graph, loop closures found by learned image retrieval, verified with sparse feature matching between two views, and lifted to metric scale with a monocular depth prior or multi-view transformer. However, 2GO handles one session at a time. It cannot relate independent sessions to each other, and its consistency check relies on a live trajectory estimate that exists from the first keyframe onward, which a new session does not have.

This paper presents Polygraph, a multi-session trajectory optimization pipeline that closes this gap. Polygraph processes each session online, one keyframe and one spatial cluster at a time. It does not assume that all sessions are available up front, and it does not fix any session as a reference. When a cluster closes, it is matched with every other session by learned retrieval. Candidate pairs are filtered and verified with sparse matching, monocular depth, and robust pose estimation, and reduced to one confirmed match per cluster pair. A cross-session edge only enters the shared graph once a two-tier consensus gate admits it, which protects the graph from spurious matches. A cheap incremental solver keeps the graph current after every cluster, and a periodic batch solve prunes outlier edges. The algorithm is agnostic to the source of the odometry measurements and simply takes calibrated images and odometry edges as input.

We evaluate Polygraph on stereo outdoor driving on KITTI~\cite{geiger2012kitti}, HoloLens and iPad indoor traversals from LaMAR~\cite{sarlin2022lamar}, legged indoor locomotion from CroCoDL~\cite{blum2025crocodl}, vehicle and handheld routes from the large-scale outdoor VBR dataset~\cite{vbr} and small-scale indoor EuRoC dataset~\cite{burri2016euroc}. Against the closest prior work 2GO on the same GPU model, using its reported timings, Polygraph runs faster, commits fewer loop-closure edges, and
matches or improves on accuracy.

In summary, this paper makes the following contributions.
\begin{itemize}
  \item An online, reconstruction-free formulation of multi-session pose graph optimization that needs neither all sessions up front nor a fixed reference session.
  \item A cluster-level matching and joint pose estimation pipeline with a two-tier consensus gate that admits cross-session edges only once they are corroborated.
  \item A characterization of runtime, accuracy, and loop-closure accuracy across a wide range of devices and scenes, including hardware-matched comparison against 2GO and comparison against other similar methods.
\end{itemize}

\section{RELATED WORK}

\subsection{Large-Scale and Multi-Session SLAM}

ORB-SLAM3~\cite{campos2021orbslam3} and VINS-Fusion~\cite{qin2019vinsfusionlocal,qin2019vinsfusionglobal} track a camera in real time and reach global consistency through loop closures and bundle adjustment over a landmark map, which scales poorly with map size. Multi-session systems extend this recipe. ORB-SLAM3 Atlas~\cite{elvira2019orbslamatlasrobustaccuratemultimap} merges several maps on place recognition, Maplab~\cite{schneider2018maplab,cramariuc2023maplab2} offers an open multi-session visual-inertial framework, and COVINS~\cite{covins} adds a centralized back-end on an ORB-SLAM3 front-end. All rely on map points, so fusion cost is tied to keeping geometry per session. Polygraph uses such a system, or a device's own odometry, only as input. Its cost scales with keyframes rather than scene density, and its online solve builds on iSAM2~\cite{kaess2012isam2}.

\subsection{Visual Place Recognition}

Deciding which keyframes might show the same place is a retrieval problem, with each image embedded into a global descriptor for nearest-neighbor search. NetVLAD~\cite{arandjelovic2016netvlad} introduced the learned aggregation approach, and recent descriptors such as SALAD~\cite{izquierdo2024salad}, BoQ~\cite{alibey2024boq}, and MegaLoc~\cite{berton2025megaloc} improve robustness to viewpoint and appearance changes. Polygraph uses such a descriptor to cast votes between clusters of keyframes. Unlike single-session loop closure, its index spans every other session and must grow as new sessions arrive online.

\subsection{Learned Two-View Matching and Monocular Priors}

A retrieval hit is only a hypothesis and must be verified with correspondences between the two images. Learned matchers such as SuperPoint~\cite{detone2018superpoint} with SuperGlue~\cite{sarlin2020superglue}, DISK~\cite{tyszkiewicz2020disk} with LightGlue~\cite{lindenberger2023lightglue}, the lightweight XFeat~\cite{potje2024xfeat}, and the detector-free LoFTR~\cite{loftr} have made two-view matching robust under large viewpoint change, but two-view geometry from correspondences alone only yields a relative pose up to scale. Dense reconstruction with MASt3R~\cite{leroy2024mast3r} instead regresses a metric point map directly from the pair, recovering scale without any external depth prior. Monocular metric depth models such as Metric3D~v2~\cite{hu2024metric3dv2} recover that same scale from a single image, independent of any two-view matcher.

\subsection{Map-Less Multi-Session Pose Graph Optimization}

Multi-robot SLAM has long treated inter-robot loop closures as pose graph edges rather than map merges. DOOR-SLAM~\cite{lajoie2020doorslam}, Kimera-Multi~\cite{tian2022kimeramulti}, and Swarm-SLAM~\cite{lajoie2024swarmslam} detect such edges with global descriptors and local feature verification, rejecting outliers with pairwise consistency maximization~\cite{mangelson2018pcm} or graduated non-convexity~\cite{yang2020gnc} before a distributed solve. FLOCC-SLAM~\cite{lajoie2025flocc} keeps that formulation but swaps classical feature verification for a 3D foundation model, closer to Polygraph's own two-view verification. COVINS-G~\cite{covins-g} centralizes the idea instead, keeping only 2D keypoints and odometry poses on the server and recovering a metric pose by treating a keyframe and its odometry neighbors as one multi-camera rig. All of these rely on a second view for scale and on every robot being known up front. 2GO~\cite{lim2025twogo} instead combines learned matching and a monocular depth prior into a pose graph pipeline that never reconstructs the scene, checking each candidate against the trajectory's own live pose estimate, always available since the graph is one connected chain from the first keyframe. Polygraph reuses this verification recipe with faster components, XFeat with LighterGlue~\cite{potje2024xfeat,lindenberger2023lightglue} and MegaLoc~\cite{berton2025megaloc}, but a newly arriving session has no live estimate and no known relationship to the rest of the graph. Its two-tier consensus gate bootstraps that trust from scratch, playing the role pairwise consistency plays in multi-robot SLAM, but at the cluster level and without assuming the set of sessions is known. Polygraph's final solve also uses graduated non-convexity to reject edges that survive gating but remain outliers.

\vspace{-5pt}

\section{METHOD}

\begin{figure*}[t]
  \centering
  \includegraphics[width=0.96\textwidth]{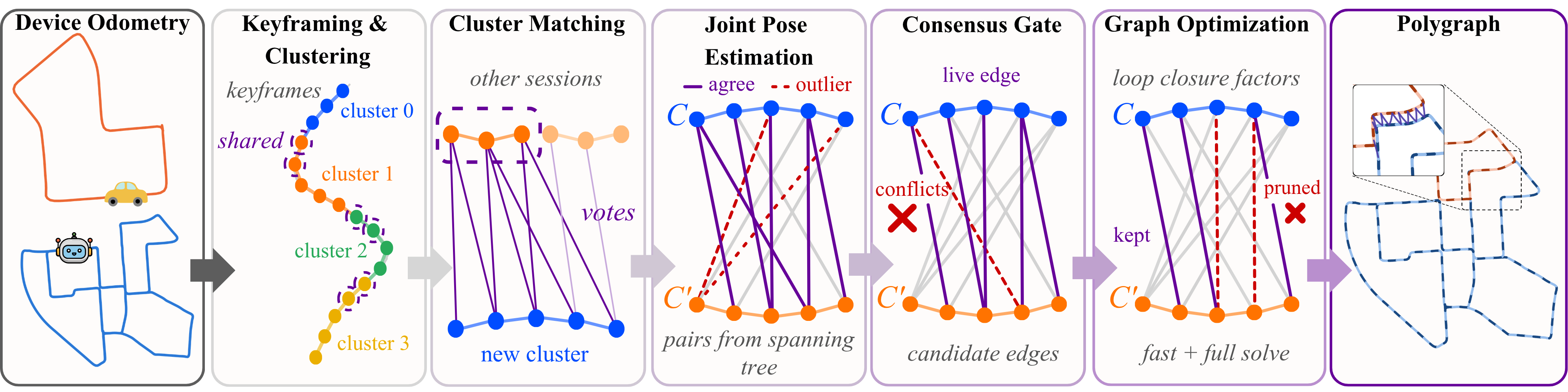}
  \caption{\textbf{Polygraph pipeline}. Each session's odometry is keyframed and grouped online into spatial clusters. A closed cluster is matched against every other session by image retrieval, candidate pairs are verified by joint pose estimation into one confirmed match, which enters the shared pose graph once the consensus gate admits it. A fast incremental solve runs after every closed cluster, and a periodic full solve prunes outliers.}
  \label{fig:pipeline}
  \vspace{-20pt}
\end{figure*}

We will start with having a brief outlook on our pipeline. Fig.~\ref{fig:pipeline} gives this overview. The odometry of each session is keyframed and clustered online. A closed cluster is matched against every other session by retrieval, candidate pairs are verified as confirmed matches, and a gate decides whether a cross-session match may enter the shared pose graph. The rest of this section follows this order.

\subsection{Problem Formulation}

Let $S$ be a set of sessions that grows over time. Each session $s \in S$ is tracked by its own odometry, which yields poses $T^{\mathrm{raw}}_{s,i} \in \mathrm{SE}(3)$ in a local frame that drifts and has no known relation to any other session. Polygraph thins this stream to keyframes and works only at the keyframe level from then on. We write $T_{s,j}$ for the pose of the $j$-th keyframe of session $s$.

We build a graph $G = (V, E)$ with one vertex per keyframe. The edge set has three parts. Odometry edges connect consecutive keyframes of one session with the relative pose read directly from the odometry, trusted and never removed. Intra-session edges are loop closures between non-consecutive keyframes of the same session; cross-session edges connect keyframes of two different sessions, and both types come from the same retrieval and verification pipeline. Only cross-session edges pass through the consensus gate, since only they can move one session relative to another. With $X_v$ the pose of vertex $v$ in the shared frame of its connected component, the graph is solved by
\begin{equation}
  \label{eq:pgo}
  \min_{\{X_v\}} \sum_{(u,v) \in E} \rho_{uv}\big( r(X_u, X_v, z_{uv}) \big) + \sum_{v \in V_0} \| X_v \ominus p_v \|^2_{\Sigma_0},
\end{equation}
where $r(X_u, X_v, z) = \mathrm{Log}(z^{-1} X_u^{-1} X_v)$ is the residual $\mathrm{SE}(3)$ of the measurement $z_{uv}$ under its covariance, $X_v \ominus p_v = \mathrm{Log}(p_v^{-1} X_v)$ is the same tangent-space difference against the prior $p_v$, and $\rho_{uv}$ is the identity for the edges of the odometry and a Cauchy loss for loop closures. The set $V_0$ contains one gauge-fix prior per session. Two properties separate this from an offline formulation. Both $V$ and $E$ grow one keyframe and edge at a time, and $V_0$ is never fixed in advance. Every session starts as its own component with its own prior, and components merge when a cross-session edge survives the gate.

\subsection{Keyframing and Online Clustering}

A raw frame becomes a keyframe once it has moved more than $\tau_{\mathrm{t}}$ or rotated more than $\tau_{\mathrm{r}}$ from the previous keyframe. Each keyframe receives a MegaLoc~\cite{berton2025megaloc} global descriptor only at that time. The keyframes are then grouped into spatially bounded clusters using only the session's own travel distance, as in the second panel of Fig.~\ref{fig:pipeline}. A cluster closes once it has at least $n_{\min}$ keyframes and has traveled more than $d_{\max}$ from its first keyframe or reached $n_{\max}$ keyframes. The last $n_{\mathrm{ov}}$ keyframes of a closing cluster seed the next one, so consecutive clusters chain without a coverage gap. Clusters, not keyframes, are the unit that all later stages discuss. A cluster becomes visible only to the rest of the pipeline once it closes.

\subsection{Cluster Matching}
We identify loop-closure candidates by matching clusters, by checking potential matches of a closed cluster $C$ against every other cluster. To reduce computational cost, we first apply a cheap retrieval filter that reduces the field to a short list of candidate pairs. It runs once per session $s'$, including $C$'s own session for intra-session loop closures, and never ranks candidates from different sessions against each other. Every keyframe of $C$ queries the descriptors of $s'$ by cosine similarity, as sketched in the cluster matching panel of Fig.~\ref{fig:pipeline}, and casts one vote for the parent cluster of each of its $k_{\mathrm{top}}$ nearest neighbours whose similarity clears a floor. When $s'$ is the session of $C$ itself, the two clusters adjacent to $C$ are excluded, since they share keyframes by construction. A session in which fewer than $n_{\mathrm{part}}$ keyframes of $C$ found anything to vote for is dropped. Otherwise, each candidate $C'$ is scored by its vote share
\begin{equation}
  \label{eq:share}
  \mathrm{share}(C, C') = \frac{\text{votes for } C' \text{ within } s'}{\text{all votes cast within } s'},
\end{equation}
Candidates with a share greater than $\tau_{\mathrm{share}}$ are kept, at most $k_{\mathrm{cl}}$ per session. The output is a short list of candidates for cluster pairs, not a list of matches. Proximity retrieval, as in~\cite{lim2025twogo}, adds a second, appearance-free candidate source within the same session, using odometry position and heading, not descriptors. Proximity retrieval is by default turned off, but as we will see later, it can be helpful when dealing with long loop-less trajectories. The candidates for loop-closure cluster pairs are then checked for a robust consensus in estimated relative pose.

\begin{figure}[h]
  \centering
  \includegraphics[width=\columnwidth]{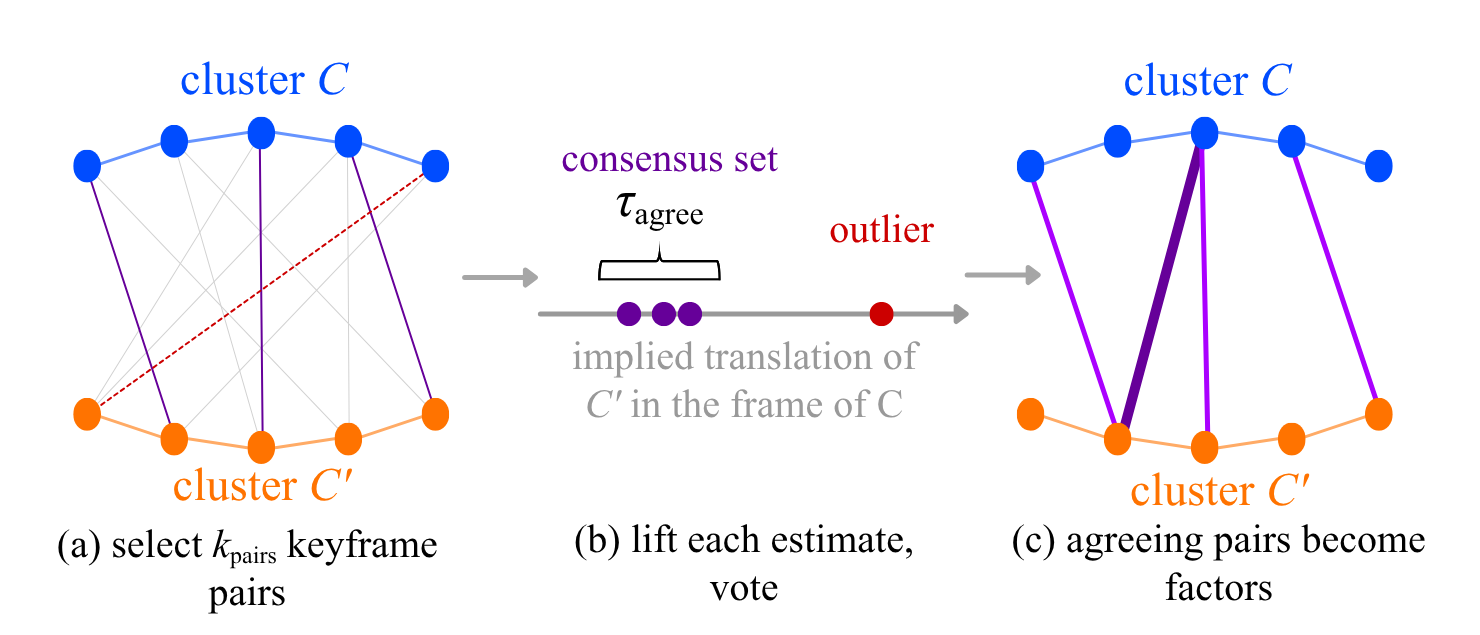}
  \caption{\textbf{Joint pose estimation}. Turns one candidate cluster pair into one confirmed match. A spanning tree over descriptor similarity selects a few keyframe pairs (purple, one wrong pair in red). Each pair yields its own metric pose, which is lifted through the odometry into the common cluster frame. Estimates that agree within $\tau_{\mathrm{agree}}$ form the consensus set, and the outlier is dropped. The chordal mean of the set is the measurement the gate decides on, and the agreeing pairs enter the graph as factors once the match is admitted.}
  \label{fig:cluster_edge}
  \vspace{-5pt}
\end{figure}

\subsection{Joint Pose Estimation}
This stage is the only one that looks at image content, and it turns one candidate pair into one confirmed match or rejects it, as Fig.~\ref{fig:cluster_edge} illustrates. Relative pose estimation between each keyframe of $C$ and every keyframe of $C'$ would be redundant, so a maximum spanning tree over the bipartite descriptor-similarity graph selects $k_{\mathrm{pairs}}$ keyframe pairs that are both similar and spread across both clusters.

Each selected pair $(k, k')$ goes through the same chain that closely follows~\cite{lim2025twogo}: the features of XFeat~\cite{potje2024xfeat} are matched with LighterGlue~\cite{potje2024xfeat,lindenberger2023lightglue}. Essential-matrix RANSAC yields a scale-free relative rotation and translation direction with an inlier set, and rejects the pair early if too few inliers survive. Inliers are triangulated into scale-free 3D points in the frame of $k$. Metric3D~v2 ~ \cite{hu2024metric3dv2} predicts a metric depth map for each image independently, and every matched keypoint is back-projected through its own depth map to a metric point. 
For each correspondence, one scalar is fit to best align its 3D triangulated point with its 3D depth-lifted point, an overdetermined one-parameter fit across x, y, and z, and correspondences whose residual after scaling exceeds $\tau_{\mathrm{scale}}$ are discarded. Finally, the surviving depth-lifted points of $k'$ and their pixel observations in $k$ are handed to PnP RANSAC, which yields the metric pose $\hat{T}_{k,k'} \in \mathrm{SE}(3)$. 
A pair is rejected unless enough correspondences survive the depth and scale-consistency checks to run PnP at all, and the resulting (post-RANSAC) inlier ratio and model score clear their own floors.

To summarize transforms between keyframes into a transform between clusters, we propagate each individual estimate along the odometry edges into a relative transform between the two clusters' reference keyframes. Two estimates agree if their implied translations differ by less than $\tau_{\mathrm{agree}}$. The estimate with the most agreeing pairs anchors a consensus set, averaged by a chordal $\mathrm{SE}(3)$ mean into the measurement $\hat z$ between the two clusters' reference keyframes. A candidate becomes a confirmed match only if the consensus set holds at least $n_{\mathrm{vote}}$ estimates; a single uncorroborated estimate is never enough, and the gate and outlier checks act on $\hat z$ alone. With \texttt{multi\_edge\_consensus} on, every consensus-set member is also added as its own factor once a match is admitted, so the optimizer sees the spread of evidence rather than only its mean. This proved especially helpful for densifying the graph when revisits are sparse.

\subsection{Two-Tier Consensus Gate}

\begin{figure}[t]
  \centering
  \includegraphics[width=0.92\columnwidth]{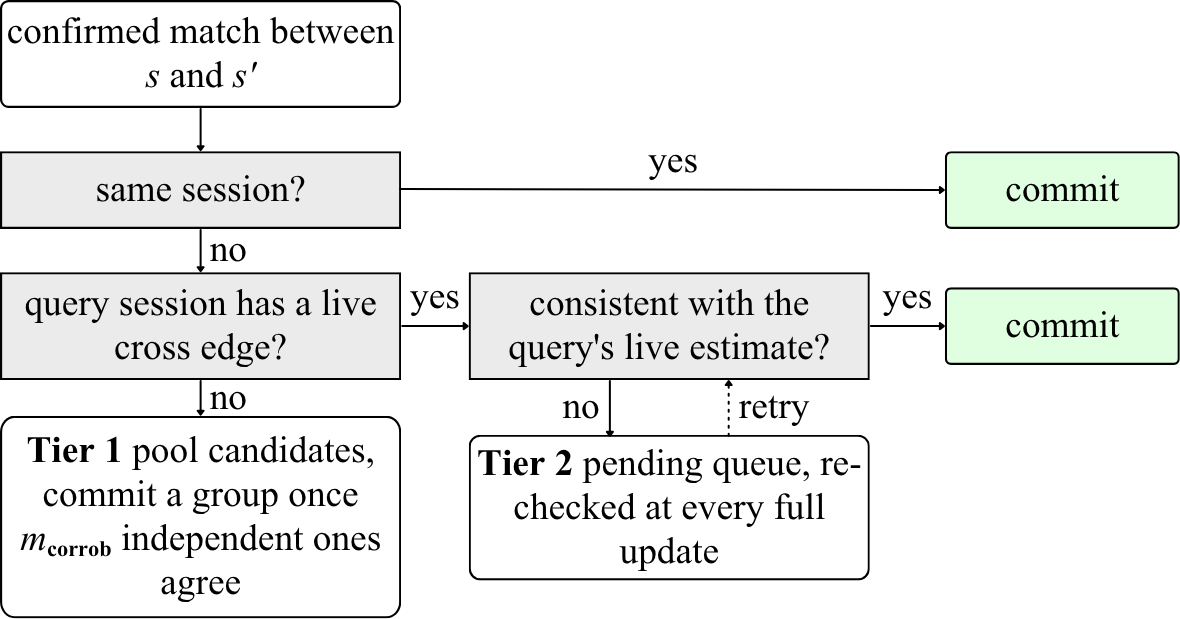}
  \caption{\textbf{The consensus gate}. Same-session matches commit directly. A cross-session edge whose query session has no live cross-session edge yet must find an agreeing, independent peer in the corroboration pool. Once a session is connected, each edge is checked against its live estimate and queued for later re-checks if it fails.}
  \label{fig:gate}
  \vspace{-5pt}
\end{figure}

A confirmed cross-session match is still a single piece of evidence from one retrieval hit, and admitting it directly would let one spurious match merge two sessions. The gate in Fig.~\ref{fig:gate} decides whether and when a match enters the graph as a loop-closure edge. An edge whose clusters both belong to the same session commits immediately, since that session's position is already fixed by its odometry chain, and the edge only adds corroboration.

For a cross-session edge between $s$ and $s'$ the gate first picks an anchor and a query. Let $n(s)$ be the number of live cross-session edges incident on $s$. The session with more live edges is the anchor, its estimate fixed for this check, and the other is the query, whose count selects a tier.

\underline{Tier 1, corroboration pool.} If the query has no live cross-session edge, the candidates that target it are pooled and checked only against each other. Once at least $m_{\mathrm{corrob}}$ pooled candidates imply mutually consistent placements, within $\tau_{\mathrm{pool}}$ in translation and rotation, and are independent (anchor-side clusters at least $g_{\min}$ apart), they commit together, so the query session always joins with two or more edges at once, never exactly one.

\underline{Tier 2, pending queue.} If the query already has a live cross-session edge, the candidate is checked immediately against its current estimate under a translation and rotation tolerance and commits if consistent. If not, it enters a pending queue, re-checked at every full update, until it passes, collects $r_{\max}$ rejections, or ages out, so an edge that arrived before the graph settled can still be admitted later.

The gate is what allows a session to arrive with no known relation to the rest of the graph. It plays the role that pairwise consistency checks play in multi-robot SLAM, but at cluster level, online, and without assuming that the set of sessions is known.

\subsection{Pose Graph Optimization}

\begin{table*}[t]
  \centering
  \footnotesize
  \setlength{\tabcolsep}{2.9pt}
  \renewcommand{\arraystretch}{1.0}
  \caption{\textbf{RMSE ATE on KITTI's loop sequences and on VBR}. Refining ORB-SLAM3 (left) and VINS-Fusion (right) odometry~\cite{lim2025twogo}. Best per odometry source shaded and bold, second best underlined.}
  \begin{tabular*}{\textwidth}{@{\extracolsep{\fill}} l r r | r r r r r r r | r r r r r r r}
    \toprule
    & & & \multicolumn{7}{c|}{ORB-SLAM3 odometry} & \multicolumn{7}{c}{VINS-Fusion odometry} \\
    Sequence & Length & Duration & odom & ORB- & Maplab & Maplab & 2GO & Ours & Ours & odom & VINS- & Maplab & Maplab & 2GO & Ours & Ours \\
    & & & only & SLAM3 & BIN & SP & & single & multi & only & PGO & BIN & SP & & single & multi \\
    \midrule
    KITTI-00 & 3.72 km & 470 s & 4.20 & 4.85 & 37.51 & 12.69 & 1.82 & \second{1.28} & \best{1.27} & 13.09 & 4.20 & 11.13 & 11.39 & 1.58 & \second{1.34} & \best{1.33} \\
    KITTI-02 & 5.07 km & 483 s & 8.05 & \second{5.64} & 18.01 & 44.78 & 8.04 & \best{5.30} & \best{5.30} & 21.02 & 13.03 & 23.96 & 18.29 & \second{10.58} & \best{7.96} & \best{7.96} \\
    KITTI-05 & 2.21 km & 288 s & 2.05 & 0.95 & 7.44 & 7.98 & \second{0.92} & \best{0.89} & \best{0.89} & 6.39 & 3.68 & 11.37 & 11.21 & \best{1.70} & \second{1.88} & \second{1.88} \\
    KITTI-06 & 1.23 km & 114 s & 2.05 & 0.79 & 7.09 & 10.74 & \best{0.62} & \second{0.72} & \second{0.72} & 3.54 & 1.68 & 13.27 & 14.76 & \second{1.48} & \best{1.36} & \best{1.36} \\
    KITTI-07 & 0.69 km & 114 s & 1.19 & 0.50 & 5.60 & 5.63 & \second{0.41} & 0.48 & \best{0.38} & 2.11 & 0.65 & 7.48 & 7.70 & \best{0.46} & 0.57 & \second{0.55} \\
    KITTI-08 & 3.22 km & 423 s & \second{3.50} & 3.67 & 40.79 & 11.65 & 3.78 & \best{3.24} & \best{3.24} & 10.43 & 10.01 & 17.04 & 14.27 & \best{4.03} & \second{8.46} & \second{8.46} \\
    KITTI-09 & 1.71 km & 165 s & 2.87 & 3.22 & 10.23 & 11.32 & \second{1.84} & \best{1.71} & \best{1.71} & 7.80 & 7.71 & 15.97 & 12.31 & \second{4.66} & \best{3.32} & \best{3.32} \\
    \textit{Total (L2)} & & & 10.62 & 9.00 & 60.31 & 51.39 & 9.33 & \second{6.69} & \best{6.68} & 28.99 & 19.07 & 40.06 & 34.96 & \second{12.56} & \best{12.39} & \best{12.39} \\
    \midrule
    campus\_train0 & 2.73 km & 602 s & 28.92 & 11.94 & 30.73 & 28.22 & \best{9.32} & \second{9.84} & 10.27 & 47.74 & 43.84 & 49.46 & 46.28 & 12.30 & \second{11.65} & \best{11.55} \\
    campus\_train1 & 2.95 km & 584 s & 28.56 & 7.99 & 30.88 & 27.42 & 8.98 & \second{6.43} & \best{6.31} & 46.94 & 23.40 & 44.43 & 34.93 & 9.60 & \second{8.75} & \best{7.82} \\
    ciampino\_train0 & 9.01 km & 1549 s & 82.82 & 68.11 & 83.89 & 81.77 & \second{18.29} & \best{18.25} & \best{18.25} & 131.33 & 95.95 & 130.20 & 130.94 & 22.84 & \best{16.72} & \best{16.72} \\
    ciampino\_train1 & 5.20 km & 942 s & 25.98 & 13.83 & 27.61 & 25.00 & \second{13.25} & \best{13.22} & \best{13.22} & 47.21 & 34.68 & 40.94 & 40.79 & \second{14.20} & \best{12.91} & \best{12.91} \\
    colosseo\_train0 & 1.45 km & 882 s & 11.09 & 9.67 & 9.40 & 10.29 & \second{6.80} & \best{4.95} & \best{4.95} & 21.49 & 13.37 & 21.48 & 21.47 & \second{6.00} & \best{4.99} & \best{4.99} \\
    diag\_train0 & 1.04 km & 1003 s & / & / & / & / & / & / & / & 27.35 & 33.10 & 27.17 & 27.23 & \second{26.70} & \best{2.97} & \best{2.97} \\
    pincio\_train0 & 1.28 km & 1114 s & 6.30 & \second{2.08} & 6.31 & 6.91 & 2.79 & \best{1.96} & \best{1.96} & 11.29 & 7.75 & 11.22 & 11.23 & \best{3.18} & \second{4.75} & \second{4.75} \\
    spagna\_train0 & 1.56 km & 1414 s & 11.42 & \best{5.62} & 11.95 & 12.01 & \second{6.48} & 7.53 & 7.53 & 21.14 & 13.22 & 20.83 & 20.35 & \second{3.73} & \best{3.16} & \best{3.16} \\
    \textit{Total (L2)} & & & 97.36 & 71.88 & 99.84 & 95.69 & 27.81 & \best{27.04} & \second{27.17} & 160.45 & 119.95 & 157.50 & 154.64 & 41.71 & \second{26.92} & \best{26.59} \\
    \bottomrule
  \end{tabular*}
  \label{tab:kitti_vbr}
  \vspace{-20pt}
\end{table*}

Every admitted edge becomes a factor in \eqref{eq:pgo}. A fast update runs iSAM2~\cite{kaess2012isam2} after every closed cluster, folding in new vertices and edges incrementally, while a full update runs every $n_{\mathrm{full}}$ closed clusters as a batch solve over the whole graph. After that, a loop-closure edge is pruned if its residual exceeds $\kappa$ times the median residual of its group; intra-session and cross-session edges are judged against their own medians so one session's tight loop closures do not raise the bar for everyone else. The full update also drains the pending queue against the fresh estimate.

Once every session has been processed, one last full update runs and two final passes follow. First, each session pair's edges are cross-checked using the pre-fusion local poses captured when each edge was created, exposing groups of edges that are wrong but mutually consistent, which residual-based pruning misses since least squares simply bends the graph to satisfy them. Second, a graduated non-convexity solve with a truncated least-squares loss~\cite{yang2020gnc} runs over the complete graph and assigns every loop closure a final outlier weight. 
The result is one pose graph over every connected session. Two sessions never connect with fewer than two corroborated loop closures at once, and once connected, they are solved jointly, so the correction propagates across each trajectory as a whole rather than staying local to the overlap.
\section{EXPERIMENTS}

\vspace{-5pt}

We evaluate on the five datasets given in Table~\ref{tab:datasets}. KITTI~\cite{geiger2012kitti} and VBR~\cite{vbr} are large outdoor benchmarks that allow us to test scalability to long trajectories and large maps. HYDRO and HGE are indoor sites captured by more than one device and test robustness to viewpoint and hardware. Lastly, EuRoC~\cite{burri2016euroc} lets us isolate pose-estimation accuracy against a two-view baseline, used only in the ablation section below. We report the following metrics for our experiments: ATE is the RMSE translation error after a per-session rigid alignment to ground truth. Where sessions share ground truth coverage, we also report reference-anchored consistency: one session per component aligned to ground truth, and that fixed transform applied to the rest. This latter metric exposes a multi-session trajectory that is locally consistent but globally wrong, in a way that ATE alone cannot.

\begin{table}[h]
\vspace{5pt}
  \centering
  \footnotesize
  \setlength{\tabcolsep}{3.5pt}
  \renewcommand{\arraystretch}{1.0}
  \caption{\textbf{Datasets used throughout Experiments}. Multi-device marks a site captured by more than one platform. VBR's multi-session sites include two campus sessions.}
  \begin{tabular*}{\columnwidth}{@{\extracolsep{\fill}} l l l r c}
    \toprule
    Dataset & Setting & Devices & Sess. & Multi-device \\
    \midrule
    KITTI & outdoor, driving & car, stereo & 7 & no \\
    VBR & driving, handheld & car / handheld & 8 & no \\
    HYDRO & indoor & Spot, iPad & 10 & yes \\
    HGE & indoor & HoloLens, iPad & 9 & yes \\
    EuRoC & indoor, MAV & MAV, stereo(-IMU) & 11 & yes \\
    \bottomrule
  \end{tabular*}
  \label{tab:datasets}
  \vspace{-2pt}
\end{table}

\subsection{Single-Session}
\label{sec:single-session}

We evaluate on KITTI and VBR. Both datasets are outdoor, large-scale, and the setting our method targets. We match the experimental setup of 2GO~\cite{lim2025twogo}, refining the same raw odometry sources in Table~\ref{tab:kitti_vbr}. `Single' treats a sequence alone, `multi' runs a dataset together with cross-session edges, and `Total' is the L2 norm over a dataset's rows. ORB-SLAM3 lost tracking on diag\_train0, excluded from that column. VBR's VINS-Fusion backbone uses one config per platform (keyframe spacing 5.0 m vehicle, 2.5 m handheld). Its ORB-SLAM3 backbone needed one per location instead, since ORB-SLAM3's larger rotation drift on VBR's tight turns is harder for retrieval and pose verification to absorb. On KITTI, each odometry shares a single configuration file across all 7 sessions. 

\begin{figure*}[t]
  \centering
  \includegraphics[width=\textwidth]{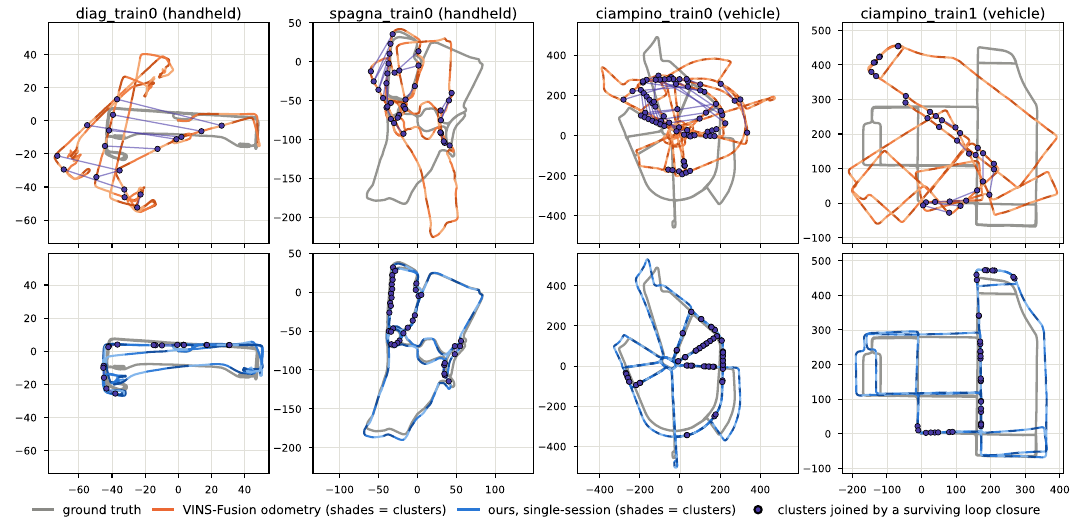}
  \caption{\textbf{Single-session VBR visualisation}. Two handheld and two vehicle. Top shows VINS-Fusion odometry; bottom shows Polygraph, both against ground truth in gray. Shades mark spatial clusters; dots mark clusters joined by a surviving loop closure.}
  \label{fig:vbr_strip}
  \vspace{-10pt}
\end{figure*}

Polygraph's single-session column is already competitive with, and usually ahead of, 2GO on both odometry sources. Fig.~\ref{fig:vbr_strip} shows four VBR routes, single-session only. Odometry drifts by tens to hundreds of meters, and Polygraph pulls every route back close to ground truth from retrieval, two-view matching, and a monocular depth prior alone, using only that session's own revisits. This evaluation indicates that the additional sparsity from keyframe clustering, compared to the direct loop-closures between keyframes in 2GO or the dense bundle-adjustment in ORB-SLAM3 does not come at a cost of accuracy, and instead yields overall more accurate single-session solutions. We attribute this accuracy gain to the additional filtering and consensus steps for the clusters.

\subsection{Multi-Session}

The multi-session results in Table~\ref{tab:kitti_vbr} on
KITTI and VBR overall show improvements compared to the single-session results. Importantly, multi-session never falls behind single-session by more than 0.03 m anywhere. However, trajectories on these datasets in general barely overlap geographically.

\PAR{Multi-Robot Split.} Table~\ref{tab:door_swarm} gives KITTI a second, genuinely multi-session test, following DOOR-SLAM~\cite{lajoie2020doorslam} and Swarm-SLAM's~\cite{lajoie2024swarmslam} own KITTI-00 split protocol, one sequence cut into 2 or 3 equal parts, each its own session on its own odometry, run concurrently. Joint ATE and RMSE after one rigid alignment of every robot together stay well below both baselines on every odometry source we tried. RTF is against the longest session in the split, both means of 5 runs.

\begin{table}[t]
\vspace{5pt}
  \centering
  \scriptsize
  \caption{\textbf{KITTI multi-robot experiment}. KITTI-00 cut into 2 or 3 equal parts, one robot each, against DOOR-SLAM and Swarm-SLAM on their own protocol, RTAB-Map stereo odometry for both. Best shaded and bold, second underlined, per split. RTF is conservative, run on a shared cluster.}
  \setlength{\tabcolsep}{3pt}
  \renewcommand{\arraystretch}{1.0}
  \begin{tabular*}{\columnwidth}{@{\extracolsep{\fill}} l c c c}
    \toprule
    Method & Split (robots) & Joint ATE (m) & RTF \\
    \midrule
    Swarm-SLAM~\cite{lajoie2024swarmslam} & 2 & 2.17 & \tbd \\
    Ours, w/ RTAB-Map odom. & 2 & \best{1.16} & 0.81 \\
    Ours, w/ ORB-SLAM3 odom. & 2 & \second{1.22} & 0.74 \\
    Ours, w/ VINS-Fusion odom. & 2 & 1.47 & 0.73 \\
    \midrule
    DOOR-SLAM~\cite{lajoie2020doorslam} & 3 & 8.00 & \tbd \\
    Ours, w/ RTAB-Map odom. & 3 & \second{1.26} & 0.95 \\
    Ours, w/ ORB-SLAM3 odom. & 3 & \best{1.20} & 0.91 \\
    Ours, w/ VINS-Fusion odom. & 3 & 2.31 & 0.97 \\
    \bottomrule
  \end{tabular*}
  \label{tab:door_swarm}
\end{table}

\PAR{HYDRO~\cite{blum2025crocodl} and HGE~\cite{sarlin2022lamar}} allow for large-scale multi-session evaluations. Both sites are captured across many sessions per device: 5 Spot and 5 iOS sessions at HYDRO, and 5 HoloLens and 4 iOS sessions at HGE, each device already tracked by onboard odometry. We tune one config per site on its primary device, Spot at HYDRO and HoloLens at HGE, then run iOS with the same config. Table~\ref{tab:hydro_hge} reports single-session and multi-session ATE per device. Multi-session beats both raw odometry and single-session on every device at both sites, even though each device's built-in odometry already achieves high accuracy. Polygraph successfully improves the poses from these odometry sources in each case.

Overall, we observe that our multi-session optimization is effective across 4 datasets and outperforms Swarm-SLAM and DOOR-SLAM on their KITTI-Splits by large margins.


\begin{table}[t]
\vspace{5pt}
  \centering
  \scriptsize
  \setlength{\tabcolsep}{1.6pt}
  \renewcommand{\arraystretch}{1.0}
  \caption{\textbf{RMSE ATE on HYDRO~\cite{blum2025crocodl} (5 Spot and 5 iOS sessions, footprint diagonal 89 m) and HGE~\cite{sarlin2022lamar} (5 HoloLens and 4 iOS sessions, footprint diagonal 167 m)}. Each device on its own odometry, and reference-anchoring error with every session of a site placed into one shared frame. Pooled per device as mean and L2 norm. Best of multi-session and mixed shaded and bold, second best underlined.}
  \begin{tabular*}{\columnwidth}{@{\extracolsep{\fill}} c l l | c c c | c c c}
    \toprule
    & & & \multicolumn{3}{c|}{ATE (m)} & \multicolumn{3}{c}{Anchoring, mixed} \\
    & Device & & odom only & multi-session & mixed & trans (m) & trans (\%) & rot (deg) \\
    \midrule
    \multirow{4}{*}{\rotatebox{90}{HYDRO}} & \multirow{2}{*}{Spot} & Mean & 0.299 & \second{0.149} & \best{0.148} & 0.181 & 0.20 & 0.39 \\
     & & L2 & 0.727 & \best{0.336} & \best{0.336} & 0.362 & 0.41 & 0.79 \\
     & \multirow{2}{*}{iOS} & Mean & 0.408 & \second{0.312} & \best{0.309} & 0.418 & 0.47 & 0.62 \\
     & & L2 & 0.990 & \second{0.722} & \best{0.721} & 0.963 & 1.08 & 1.42 \\
    \midrule
    \multirow{4}{*}{\rotatebox{90}{HGE}} & \multirow{2}{*}{HoloLens} & Mean & 0.658 & \best{0.280} & \second{0.281} & 0.956 & 0.57 & 0.84 \\
     & & L2 & 2.192 & \best{0.668} & \best{0.668} & 2.104 & 1.26 & 1.73 \\
     & \multirow{2}{*}{iOS} & Mean & 0.555 & \second{0.531} & \best{0.523} & 0.697 & 0.42 & 0.94 \\
     & & L2 & 1.190 & \second{1.131} & \best{1.121} & 1.445 & 0.87 & 2.04 \\
    \bottomrule
  \end{tabular*}
  \label{tab:hydro_hge}
\end{table}

\subsection{Multi-Device}

\begin{figure*}[t]
  \centering
  \includegraphics[width=\textwidth]{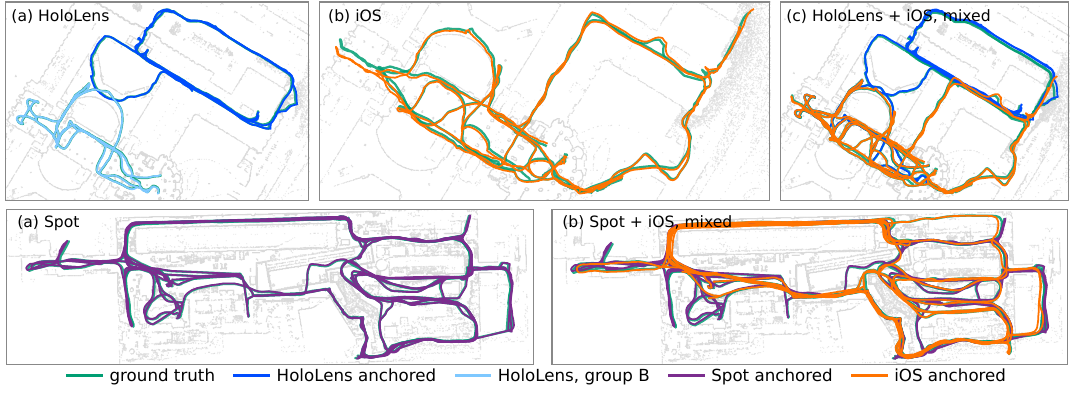}
  \vspace{-18pt}
  \caption{\textbf{Reference anchoring on HGE (top) and HYDRO (bottom)}. One fixed transform per connected component applied to every session past its reference. Ground truth in green, Polygraph's estimate on top. On HGE, (a) is HoloLens alone with two reference groups, (b) is iOS alone, (c) is HoloLens and iOS anchored together. On HYDRO, (a) is Spot alone, (b) is Spot and iOS together, all ten sessions in one component.}
  \label{fig:anchoring}
  \vspace{-10pt}
\end{figure*}

Beyond combining sessions from the same device, HYDRO and HGE let us mix two entirely different devices into one shared graph. The results for this are shown in Table~\ref{tab:hydro_hge}'s mixed column. Mixed usually beats standalone multi-session, too, since a second device adds more chances to close a loop. No external baseline exists for either site. Anchoring translation is in meters and percent of the site's footprint diagonal, allowing comparison of HYDRO and HGE despite their different scale. Fig.~\ref{fig:anchoring} shows why the reference-anchoring numbers matter beyond ATE alone. HGE's HoloLens sessions split into two disconnected reference groups on their own, and adding iOS bridges them into one shared frame.

\subsection{Ablation}

\PAR{Camera Pairing.} Table~\ref{tab:ablation_cameras} shows that sensor rigs with more cameras help most where they add a genuinely new viewpoint. HYDRO's four cameras (from Spot) more than double cross-session edges and improve single-session ATE, while KITTI's second camera, the other half of the same stereo rig, changes nothing.

\begin{table}[t]
\vspace{5pt}
  \centering
  \scriptsize
  \setlength{\tabcolsep}{1.3pt}
  \renewcommand{\arraystretch}{1.0}
  \caption{\textbf{Camera-pairing ablation}. Baseline is single-camera production; best is the best combination available. The better ATE per pair is shaded and bold.}
  \begin{tabular*}{\columnwidth}{@{\extracolsep{\fill}} l l r r r}
    \toprule
    Dataset & Cameras & Single (m) & Multi (m) & Cross edges \\
    \midrule
    \multirow{2}{*}{HGE} & 1, baseline & 0.652 & 0.280 & 252 \\
     & 4, best & 0.652 & \best{0.278} & 506 \\
\midrule
    \multirow{2}{*}{HYDRO} & 1, baseline & 0.236 & 0.149 & 1484 \\
     & 4, best & \best{0.197} & \best{0.146} & 3672 \\
\midrule
    \multirow{2}{*}{KITTI} & 1, baseline & 1.946 & 1.930 & 18 \\
     & 2, best & 1.946 & 1.930 & 18 \\
    \bottomrule
  \end{tabular*}
  \label{tab:ablation_cameras}
\end{table}

\begin{figure}[tb]
  \centering
  \includegraphics[width=\columnwidth]{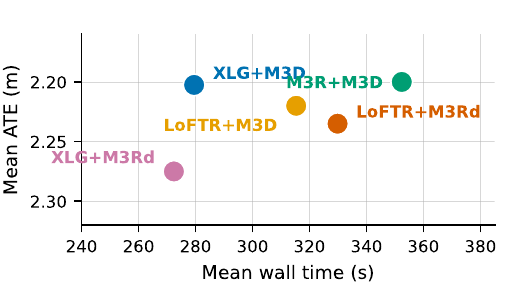}
  \caption{\textbf{Pose-estimation ablation}. Mean wall time against mean ATE across KITTI, HGE, HGE iOS, and HYDRO, both raw values, so KITTI's larger gaps keep their real weight instead of being normalized down to HGE's. XLG = XFeat+LighterGlue (production), M3R = MASt3R, M3D = Metric3D depth, M3Rd = MASt3R's own depth. XLG+M3D ties M3R+M3D on accuracy (2.20 m each) at 21\% less wall time. XLG+M3Rd is faster still but 0.07 m worse, mostly one 0.3 m KITTI gap. MASt3R-native is off this plot at 4.26 m, nearly twice everyone else's.}
  \label{fig:pe_ablation}
  \vspace{-5pt}
\end{figure}

\PAR{Pose Estimation.} Fig.~\ref{fig:pe_ablation} justifies our choice of XFeat+LighterGlue for pose estimation over LoFTR and MASt3R. Raw pair success rate does not predict full-pipeline accuracy. On HGE, MASt3R succeeds on 91.8\% of pairs against XFeat+LighterGlue's 76.8\%, yet its mean translation error on successful pairs is almost twice as high, 0.961 m against 0.544 m. The full pipeline flips this ranking. LoFTR and MASt3R-native each land sessions in badly misaligned components, and switching matcher costs 15 to 25\% more wall time for no accuracy gain. Reference anchoring makes the case more clearly than ATE. MASt3R-depth variants look competitive on raw HGE ATE but land 18 to 78 m off once anchored, a locally corroborated but globally wrong edge, the real reason production keeps Metric3D. Averaged across all four datasets, XFeat+LighterGlue with Metric3D ties MASt3R with Metric3D on accuracy, 2.20 m mean ATE each, at 21\% less wall time. Swapping only the depth source, XFeat+LighterGlue with MASt3R's own depth, is faster still but 0.07 m worse, mostly a 0.3 m gap on KITTI. Table~\ref{tab:euroc} isolates pose-estimation accuracy directly on EuRoC~\cite{burri2016euroc} versus more traditional pose estimation methods. Against COVINS-G~\cite{covins-g}, a genuinely two-view, IMU-fused system, our committed edges average 0.48 m translation error before graph optimization, compared with its 17-point variant at 0.07 m.

\begin{table}[t]
\vspace{0.5pt}
  \centering
  \scriptsize
  \setlength{\tabcolsep}{1.7pt}
  \renewcommand{\arraystretch}{1.0}
  \caption{\textbf{EuRoC~\cite{burri2016euroc}}. Accuracy of committed cross-session edges on MH01--05 before graph optimization, mean and standard deviation against ground truth, and COVINS-G values from their paper~\cite{covins-g}.}
  \begin{tabular*}{\columnwidth}{@{\extracolsep{\fill}} l l r r r}
    \toprule
    Method & Front-end & Edges & Trans. (m) & Rot. (deg) \\
    \midrule
    COVINS-G PnP & VINS-Mono & 852 & 0.69 $\pm$ 0.60 & 3.44 $\pm$ 3.5 \\
    COVINS-G 17-point & VINS-Mono & 1823 & 0.10 $\pm$ 0.17 & 1.09 $\pm$ 1.25 \\
    COVINS-G PnP & ORB-SLAM3 & 505 & 0.23 $\pm$ 0.32 & 1.56 $\pm$ 3.7 \\
    COVINS-G 17-point & ORB-SLAM3 & 406 & 0.07 $\pm$ 0.12 & 1.01 $\pm$ 0.9 \\
    Ours, stereo-inertial & ORB-SLAM3 & 43 & 0.48 $\pm$ 0.29 & 1.61 $\pm$ 0.96 \\
    Ours, stereo & ORB-SLAM3 & 62 & 0.57 $\pm$ 0.41 & 2.31 $\pm$ 1.71 \\
    \bottomrule
  \end{tabular*}
  \label{tab:euroc}
\end{table}

\PAR{Retrieval.} Retrieval is generally robust to substitution; swapping MegaLoc for BoQ, NetVLAD, or SALAD changes ATE only by a few percent.

\begin{figure}[t]
  \vspace{-10pt}
  \centering
  \includegraphics[width=\columnwidth]{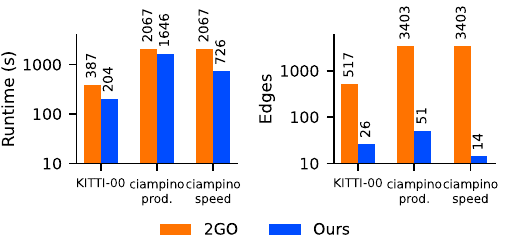}
  \vspace{-15pt}
  \caption{\textbf{Runtime and committed edges against 2GO}. Ran on the same RTX 4090, VINS-Fusion odometry, log scale. 2GO values are from their paper; ours is the mean of five runs. ciampino's speed config trims \texttt{top\_k\_clusters}/\texttt{min\_kf\_pairs} vs.\ production.}
  \label{fig:timing}
  \vspace{-5pt}
\end{figure}

\PAR{Timing.} Fig.~\ref{fig:timing} compares wall time against 2GO on the same RTX 4090. Polygraph is 1.9$\times$ faster on KITTI-00 and 1.3$\times$ faster on ciampino\_train0 in production, and 2.85$\times$ at speed, with far fewer edges: 26 and 51 (14 at speed) against 517 and 3403. Real-time factor, wall time divided by the recording's own length, is 0.44 on KITTI-00 and 0.47 at speed on ciampino, both faster than the sensor itself, and 1.06 in production. This speedup can mostly be attributed to the consensus gate rejecting single, uncorroborated matches.

Fig.~\ref{fig:pe_ablation}'s pose-estimation ablation gives the same picture, but differently. Each variant runs once per dataset, over every session it has, timed end-to-end. Dividing that wall time by the dataset's longest single session gives a real-time factor against just watching its longest recording once. Averaged across KITTI, HGE, HGE iOS, and HYDRO, this is 0.61 for production, 0.70 for MASt3R with Metric3D depth, 0.65 for LoFTR with Metric3D depth, and 0.49 for MASt3R-native, which skips the depth-lifting stage entirely and is fastest here for exactly that reason.

\PAR{Odometry Consistency.} Our pipeline is deterministic; however, odometry might not be. Therefore, we run our method end-to-end on five different realizations of ORB-SLAM3 odometry, yielding Total L2 between 5.97 and 6.63 m, close to Table~\ref{tab:kitti_vbr}'s single realization, confirming the robustness of our approach.

\PAR{Gate.} The two-tier gate is worth ablating directly. In Table~\ref{tab:gate_ablation}, we ablate admitting every confirmed match immediately with no corroboration or consistency check. Gate-off runs 5.7 to 10.3\% slower, since committing far more edges costs more downstream cleanup than the gate itself saves. As discussed above, the gate not only achieves speedup but also increases accuracy, with per-edge translation error 2.5$\times$ higher on average and rotation error almost 6$\times$ higher. In short, the gate guards against single bad edges that are hard to filter out in the optimization and therefore increases robustness and accuracy.

\begin{table}[h]
\vspace{5pt}
  \centering
  \scriptsize
  \setlength{\tabcolsep}{3pt}
  \renewcommand{\arraystretch}{1.0}
  \caption{\textbf{Two-tier consensus gate ablation}. Six HGE/HYDRO multi-session conditions pooled by dataset. Submitted are distinct cross-session candidates the gate admitted; trans/rot err is per committed edge before graph optimization. Exec $\Delta$ is gate-off vs.\ on wall time, run on an RTX 2080, not comparable in absolute terms to the pinned-4090 numbers elsewhere. Gate-on shaded and bold.}
  \begin{tabular*}{\columnwidth}{@{\extracolsep{\fill}} l r r r r r r r}
    \toprule
    & \multicolumn{2}{c}{Submitted} & \multicolumn{2}{c}{Trans err (m)} & \multicolumn{2}{c}{Rot err (deg)} & \\
    Dataset & On & Off & On & Off & On & Off & Exec $\Delta$ \\
    \midrule
    HGE & \best{250} & 373 & \best{0.399} & 2.574 & \best{0.87} & 12.41 & +7.0\% \\
    HYDRO & \best{775} & 839 & \best{0.726} & 1.223 & \best{0.90} & 1.78 & +8.6\% \\
    \textbf{Total} & \best{1025} & 1212 & \best{0.646} & 1.639 & \best{0.89} & 5.05 & +8.2\% \\
    \bottomrule
  \end{tabular*}
  \label{tab:gate_ablation}
\end{table}

\vspace{-3pt}
\section{CONCLUSION}
\vspace{-3pt}

Polygraph merges sessions online, at cluster level, using only image retrieval, two-view matching, and a monocular depth prior, without ever building or maintaining a map. Across outdoor driving, indoor multi-device capture, and a micro-aerial benchmark, it stays competitive with existing multi-session baselines while committing far fewer edges and running faster than the closest prior work. The same production config transfers across KITTI's cars, HYDRO's legged robot and phone, and HGE's headset and phone, without retuning per device. Future work includes closing the edge-count gap without giving up the accuracy it buys, and testing beyond visual and visual-inertial odometry.
\vspace{-0.15mm}

\vspace{-3pt}
\section*{\small ACKNOWLEDGEMENTS}
\vspace{-3pt}

{\small
The author(s) used generative AI to assist with drafting and editing
the manuscript, cleaning up figures and coding. All content was reviewed by the authors, who take
full responsibility for it.
\par}

\vspace{-5pt}

\bibliographystyle{IEEEtran}
\bibliography{references}

@article{hu2024metric3dv2,
  title={Metric3D v2: A Versatile Monocular Geometric Foundation Model for Zero-shot Metric Depth and Surface Normal Estimation},
  author={Hu, Mu and Yin, Wei and Zhang, Chi and Cai, Zhipeng and Long, Xiaoxiao and Chen, Hao and others},
  journal={IEEE TPAMI},
  year={2024}
}

@inproceedings{tyszkiewicz2020disk,
  title={{DISK}: Learning Local Features with Policy Gradient},
  author={Tyszkiewicz, Micha{\l} J. and Fua, Pascal and Trulls, Eduard},
  booktitle={NeurIPS},
  year={2020}
}

@inproceedings{loftr,
  title={{LoFTR}: Detector-Free Local Feature Matching with Transformers},
  author={Sun, Jiaming and Shen, Zehong and Wang, Yuang and Bao, Hujun and Zhou, Xiaowei},
  booktitle={CVPR},
  year={2021}
}

@article{elvira2019orbslamatlasrobustaccuratemultimap,
  title={{ORBSLAM-Atlas}: A Robust and Accurate Multi-Map System},
  author={Elvira, Richard and Tard{\'o}s, Juan D. and Montiel, J. M. M.},
  journal={arXiv:1908.11585},
  year={2019}
}

@article{vbr,
  title={{VBR}: A Vision Benchmark in Rome},
  author={Brizi, Leonardo and Giacomini, Emanuele and Di Giammarino, Luca and Ferrari, Simone and Salem, Omar and De Rebotti, Lorenzo and others},
  journal={arXiv:2404.11322},
  year={2024}
}

@article{covins,
  title={{COVINS}: Visual-Inertial {SLAM} for Centralized Collaboration},
  author={Schmuck, Patrik and Ziegler, Thomas and Karrer, Marco and Perraudin, Jonathan and Chli, Margarita},
  journal={arXiv:2108.05756},
  year={2021}
}

@inproceedings{covins-g,
  title={{COVINS-G}: A Generic Back-end for Collaborative Visual-Inertial {SLAM}},
  author={Patel, Manthan and others},
  booktitle={ICRA},
  year={2023}
}

@article{schneider2018maplab,
  title={maplab: An Open Framework for Research in Visual-Inertial Mapping and Localization},
  author={Schneider, Thomas and Dymczyk, Marcin and Fehr, Marius and Egger, Kevin and Lynen, Simon and Gilitschenski, Igor and others},
  journal={IEEE RA-L},
  year={2018}
}

@inproceedings{detone2018superpoint,
  title={{SuperPoint}: Self-Supervised Interest Point Detection and Description},
  author={DeTone, Daniel and others},
  booktitle={CVPRW},
  year={2018}
}

@inproceedings{sarlin2020superglue,
  title={{SuperGlue}: Learning Feature Matching with Graph Neural Networks},
  author={Sarlin, Paul-Edouard and DeTone, Daniel and Malisiewicz, Tomasz and Rabinovich, Andrew},
  booktitle={CVPR},
  year={2020}
}

@inproceedings{leroy2024mast3r,
  title={Grounding Image Matching in {3D} with {MASt3R}},
  author={Leroy, Vincent and Cabon, Yohann and Revaud, Jerome},
  booktitle={ECCV},
  year={2024}
}

@article{lim2025twogo,
  title={Loop Closure from Two Views: Revisiting {PGO} for Scalable Trajectory Estimation through Monocular Priors},
  author={Lim, Tian Yi and Sun, Boyang and Pollefeys, Marc and Blum, Hermann},
  journal={arXiv:2503.16275},
  year={2025}
}

@article{campos2021orbslam3,
  title={{ORB-SLAM3}: An Accurate Open-Source Library for Visual, Visual-Inertial, and Multi-Map {SLAM}},
  author={Campos, Carlos and Elvira, Richard and Rodr{\'\i}guez, Juan J. G{\'o}mez and Montiel, Jos{\'e} M. M. and Tard{\'o}s, Juan D.},
  journal={IEEE T-RO},
  year={2021}
}

@article{qin2019vinsfusionglobal,
  title={A General Optimization-based Framework for Global Pose Estimation with Multiple Sensors},
  author={Qin, Tong and Cao, Shaozu and Pan, Jie and Shen, Shaojie},
  journal={arXiv:1901.03642},
  year={2019}
}

@article{qin2019vinsfusionlocal,
  title={A General Optimization-based Framework for Local Odometry Estimation with Multiple Sensors},
  author={Qin, Tong and Pan, Jie and Cao, Shaozu and Shen, Shaojie},
  journal={arXiv:1901.03638},
  year={2019}
}

@article{kaess2012isam2,
  title={{iSAM2}: Incremental Smoothing and Mapping Using the {Bayes} Tree},
  author={Kaess, Michael and others},
  journal={IJRR},
  year={2012}
}

@inproceedings{arandjelovic2016netvlad,
  title={{NetVLAD}: {CNN} Architecture for Weakly Supervised Place Recognition},
  author={Arandjelovi{\'c}, Relja and Gronat, Petr and Torii, Akihiko and Pajdla, Tomas and Sivic, Josef},
  booktitle={CVPR},
  year={2016}
}

@inproceedings{izquierdo2024salad,
  title={Optimal Transport Aggregation for Visual Place Recognition},
  author={Izquierdo, Sergio and Civera, Javier},
  booktitle={CVPR},
  year={2024}
}

@inproceedings{alibey2024boq,
  title={{BoQ}: A Place is Worth a Bag of Learnable Queries},
  author={Ali-bey, Amar and Chaib-draa, Brahim and Gigu{\`e}re, Philippe},
  booktitle={CVPR},
  year={2024}
}

@inproceedings{berton2025megaloc,
  title={{MegaLoc}: One Retrieval to Place Them All},
  author={Berton, Gabriele and Masone, Carlo},
  booktitle={CVPRW},
  year={2025}
}

@inproceedings{potje2024xfeat,
  title={{XFeat}: Accelerated Features for Lightweight Image Matching},
  author={Potje, Guilherme and others},
  booktitle={CVPR},
  year={2024}
}

@inproceedings{lindenberger2023lightglue,
  title={{LightGlue}: Local Feature Matching at Light Speed},
  author={Lindenberger, Philipp and others},
  booktitle={ICCV},
  year={2023}
}

@article{yang2020gnc,
  title={Graduated Non-Convexity for Robust Spatial Perception: From Non-Minimal Solvers to Global Outlier Rejection},
  author={Yang, Heng and Antonante, Pasquale and Tzoumas, Vasileios and Carlone, Luca},
  journal={IEEE RA-L},
  year={2020}
}

@inproceedings{geiger2012kitti,
  title={Are We Ready for Autonomous Driving? {The KITTI} Vision Benchmark Suite},
  author={Geiger, Andreas and Lenz, Philip and Urtasun, Raquel},
  booktitle={CVPR},
  year={2012}
}

@article{burri2016euroc,
  title={The {EuRoC} Micro Aerial Vehicle Datasets},
  author={Burri, Michael and Nikolic, Janosch and Gohl, Pascal and Schneider, Thomas and Rehder, Joern and Omari, Sammy and others},
  journal={IJRR},
  year={2016}
}

@inproceedings{sarlin2022lamar,
  title={{LaMAR}: Benchmarking Localization and Mapping for Augmented Reality},
  author={Sarlin, Paul-Edouard and others},
  booktitle={ECCV},
  year={2022}
}

@inproceedings{blum2025crocodl,
  title={{CroCoDL}: Cross-device Collaborative Dataset for Localization},
  author={Blum, Hermann and Mercurio, Alessandro and O'Reilly, Joshua and Engelbracht, Tim and Dusmanu, Mihai and Pollefeys, Marc and Bauer, Zuria},
  booktitle={CVPR},
  year={2025}
}

@inproceedings{mangelson2018pcm,
  title={Pairwise Consistent Measurement Set Maximization for Robust Multi-Robot Map Merging},
  author={Mangelson, Joshua G. and Dominic, Derrick and Eustice, Ryan M. and Vasudevan, Ram},
  booktitle={ICRA},
  year={2018}
}

@article{lajoie2020doorslam,
  title={{DOOR-SLAM}: Distributed, Online, and Outlier Resilient {SLAM} for Robotic Teams},
  author={Lajoie, Pierre-Yves and Ramtoula, Benjamin and Chang, Yun and Carlone, Luca and Beltrame, Giovanni},
  journal={IEEE RA-L},
  year={2020}
}

@article{tian2022kimeramulti,
  title={{Kimera-Multi}: Robust, Distributed, Dense Metric-Semantic {SLAM} for Multi-Robot Systems},
  author={Tian, Yulun and Chang, Yun and Herrera Arias, Fernando and Nieto-Granda, Carlos and How, Jonathan P. and Carlone, Luca},
  journal={IEEE T-RO},
  year={2022}
}

@article{lajoie2024swarmslam,
  title={{Swarm-SLAM}: Sparse Decentralized Collaborative Simultaneous Localization and Mapping Framework for Multi-Robot Systems},
  author={Lajoie, Pierre-Yves and Beltrame, Giovanni},
  journal={IEEE RA-L},
  year={2024}
}

@article{lajoie2025flocc,
  title={{FLOCC-SLAM}: 3{D} Foundation Model-Based Loop Closing for Decentralized Collaborative {SLAM}},
  author={Lajoie, Pierre-Yves and Ramtoula, Benjamin and De Martini, Daniele and Beltrame, Giovanni},
  journal={IEEE RA-L},
  year={2025}
}

@article{cramariuc2023maplab2,
  title={maplab 2.0 -- A Modular and Multi-Modal Mapping Framework},
  author={Cramariuc, Andrei and Bernreiter, Lukas and Tschopp, Florian and Fehr, Marius and Reijgwart, Victor and Nieto, Juan and others},
  journal={IEEE RA-L},
  year={2023}
}

\end{document}